\def\year{2017}\relax
\documentclass[letterpaper]{article}
\usepackage{aaai17}
\usepackage{graphicx}

\usepackage{times}
\usepackage{helvet}
\usepackage{courier}

\usepackage{courier}
\usepackage{graphicx}
\usepackage{balance}
\usepackage{subcaption}
\usepackage{graphics} 

\usepackage{amsmath}

\usepackage{subcaption}
\usepackage{multirow}
\usepackage{multicol}

\usepackage[flushleft]{threeparttable}

\usepackage[toc,page]{appendix}

\begin{document}
%
\title{Spice up Your Chat: The Intentions and Sentiment Effects of Using Emojis}

\author{%
	  Tianran Hu$^{*1}$, Han Guo$^{\dag2}$, Hao Sun$^{*3}$, Thuy-vy Thi Nguyen$^{*4}$, Jiebo Luo$^{*5}$\\
	  	$^*$University of Rochester\\
		$^\dag$Institute of Computing Technology, Chinese Academy of Sciences\\
	  \{$^1$thu, $^5$jluo\}@cs.rochester.edu, $^2$guohan@ict.ac.cn\\
	  $^3$hao\_sun@urmc.rochester.edu, $^4$thuy-vy.t.nguyen@rochester.edu
}

\maketitle
\begin{abstract}

Emojis, as a new way of conveying nonverbal cues, are widely adopted in computer-mediated communications. In this paper, first from a message sender perspective, we focus on people's motives in using four types of emojis -- positive, neutral, negative, and non-facial. We compare the willingness levels of using these emoji types for seven typical intentions that people usually apply nonverbal cues for in communication. The results of extensive statistical hypothesis tests not only report the popularities of the intentions, but also uncover the subtle differences between emoji types in terms of intended uses. Second, from a perspective of message recipients, we further study the sentiment effects of emojis, as well as their duplications, on verbal messages. Different from previous studies in emoji sentiment, we study the sentiments of emojis and their contexts as a whole. The experiment results indicate that the powers of conveying sentiment are different between four emoji types, and the sentiment effects of emojis vary in the contexts of different valences.

\end{abstract}

\section{Introduction}
Emojis (e.g. \raisebox{-0.1em}{\includegraphics[height=0.9em]{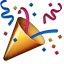}}\raisebox{-0.1em}{\includegraphics[height=0.9em]{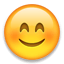}}) are defined as ``\textit{digital images that are added to messages in electronic communication}''\footnote{http://dictionary.cambridge.org/us/dictionary/english/emoji}. As effective supplements of nonverbal cues in verbal messages~\cite{lo2008nonverbal}, they are widely adopted in instant messaging, emails, social network services, and many other forms of CMC (Computer-Mediated Communication)~\cite{dresner2010functions}. It is reported that nearly half of the posts on Instagram contain emojis~\cite{dimson2015emojineering}, and emojis are replacing emoticons (e.g. ;) and :-)) on Twitter to become a popular form of representing things, feelings, concepts, and so on~\cite{pavalanathan2015emoticons}. Furthermore, studies in sociology and psychology show that emojis are creating a new language for the new generation~\cite{alshenqeeti2016emojis}. This paper explores emoji usages from two perspectives. First, from a message sender perspective, we study the intentions of using emojis in communication. Second, from a message recipient perspective, we study the effects of emojis on the sentiments of messages.

Different from face-to-face communication, CMC is lack of nonverbal cues such as face expressions, tones, and gestures, for expressing subtle information~\cite{archer1977words}. Consequently, message senders introduce many forms of surrogates for missing nonverbal cues, for example, elongated words (e.g. niiice), emoticons, and emojis. The intentions of using emoticons in CMC have been discussed in some previous work, and it is suggested that people usually use them to express sentiment, express humor, and strengthen the expression~\cite{derks2008emoticons}. However, as a newly emerged form of surrogates for nonverbal cues, the motives of using emojis are still not yet systematically studied. Although emojis resemble emoticons in some aspects, as digital graphics instead of combinations of punctuation marks, emojis are clearly more various and vivid, and reportedly more expressive than emoticons~\cite{pavalanathan2015emoticons}. Therefore, it is inappropriate to simply equate the intentions of using emojis and emoticons. Our work in this paper first focuses on the intentions of using emojis.

We summarize seven typical intentions of using nonverbal cues and their surrogates in communication, such as expressing sentiment, expressing irony, and so on. Four types of emojis are explored in this paper, and the types are positive, neutral, negative facial emojis, and a type of non-facial emojis. To study how willing people are to use the emojis for the intentions, we construct a 7 (intentions) $\times$ 20 (emojis) within-subjects user study. Extensive statistical hypothesis tests are then performed on the collected data. Test results suggest that expressing sentiment, strengthening expression, and adjusting tone are the top three most popular intentions of using emojis. Moreover, different from emoticons, people do not intend to use emojis for expressing humor. Interesting differences between different types of emojis are also discovered, for example, negative emojis are more intended to be used than positive emojis to express sentiment, and neutral emojis are the most proper for expressing irony. 

Our study in intentions reveals that expressing sentiment is the most popular intention of using emojis of message senders. A natural follow-up question is how recipients feel about the sentiments conveyed by the emojis. In previous work, the sentiments of emojis and plain verbal messages are usually discussed separately. For example, Miller et al. asked subjects to rate sentiment scores of emojis by reading plain emoji graphics~\cite{miller2016blissfully}, and Novak et al. directly took the sentiment scores of verbal messages as the scores of emojis embedded in them~\cite{novak2015sentiment}. However, the sentiments of surrogates for nonverbal cues and verbal messages are not isolated~\cite{walther2001impacts,derks2007emoticons}. A verbal message combined with different emojis could convey different sentiments. For example, the valence of message ``\textit{Just attended a seminar, now heading back home}'' is originally neutral. However, if the message was added \raisebox{-0.1em}{\includegraphics[height=0.9em]{1F60A.png}} or \raisebox{-0.1em}{\includegraphics[height=0.9em]{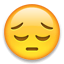}} in the end, the combinations could convey two totally different sentiments. On the other hand, the sentiments conveyed by emojis, similar to face expressions in face-to-face communication, may vary according to message contents. Two examples are ``\textit{Just got an A in the class! \raisebox{-0.1em}{\includegraphics[height=0.9em]{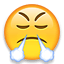}}}'', and ``\textit{My dog destroied the sofa, again... \raisebox{-0.1em}{\includegraphics[height=0.9em]{1F624.png}}}''. In the first sentence, the emoji conveys an uplifting sentiment, which is positive. The same emoji is more likely to convey a negative sentiment of anger while in the second sentence. Therefore, in this work, we study an emoji and the associated verbal message \textit{as a whole}, and discuss the sentiment effects of an emojis on the verbal messages.

We construct four separate user studies, and the sentiment effects of an emoji type is studied in each study. In the surveys, plain verbal messages of positive, neutral, and negative valences are coupled with emojis, and also with no emoji as the control conditions. Besides single emojis, it is reported that people also would like to use duplicate emojis in CMC (e.g. \raisebox{-0.1em}{\includegraphics[height=0.9em]{1F60A.png}}\raisebox{-0.1em}{\includegraphics[height=0.9em]{1F60A.png}}\raisebox{-0.1em}{\includegraphics[height=0.9em]{1F60A.png}}), and such usages are believed more powerful in expressing sentiment~\cite{tauch2016roles}. Therefore, we also add combinations of plain verbal messages and duplicate emojis in the surveys, to study their sentiment effects. We ask the participants to rate the sentiment scores for each combination of verbal messages and emojis, and statistical hypothesis tests are again conducted for data analysis. Our results reveal insights into emoji sentiment effects. For example, we discover that positive emojis do not increase the sentiment of a plain positive verbal message. Neither do negative emojis decrease the sentiment of a plain negative verbal message. Different from previous work, we find that duplicate emojis are not more powerful in expressing sentiment than single emojis in most cases.

Our work attempts to shed better light on the usage of emojis in CMC, and provides valuable insights into the sentiment effects of emojis. The main contributions of this paper are summarized as follows.

\begin{itemize}
\item We first study the intentions of using emojis in CMC from the perspective of message senders. Seven typical intentions of using nonverbal cues and their surrogates for in communication, as well as four emoji types are investigated. Our results not only explain why people employ emojis in communication, but also reveal the differences between emoji types, in terms of their most proper intentions.  

\item We then study the sentiment effects of emojis from the recipient perspective. Instead of separating emojis from verbal messages, we study the sentiment of emojis and their contexts as a whole. The results of statistical hypothesis tests suggest divergent sentiment effects of different emoji types on verbal messages of different valences.  
\end{itemize}

\section{Related Work}

\subsection{Studies on Emoji}
Previous work on emojis mainly focuses on three research directions: the meanings and sentiments of emojis, as well as the different usages of emojis among people. The technical report of Instagram is the first attempt to study the meaning of emojis using a word embedding approach~\cite{dimson2015emojineering}. In the work, the authors vectorized the emojis occurred in Instagram posts, and used the semantically closest words to the emojis in the vector space as their explanations. 
Eisner et al. also proposed an embedding model -- emoji2vec -- to learn emoji representations~\cite{eisner2016emoji2vec}. Instead of learning the emoji vectors from social media posts, they leveraged the emoji descriptions, and reported better performances on evaluation tasks. Similarly, Wijeratne et al. set up a machine readable sense inventory for emoji by aggregating the explanations of emojis from multiple online sources~\cite{wijeratne2016emojinet}. Regarding the sentiments of emojis, Novak et al built a sentiment dictionary for a large amount of emojis~\cite{novak2015sentiment}. In this work, authors asked participants to label the sentiment of messages that contain at least one emoji. The sentiment score of an emoji was then computed as the average scores of all the messages where this emoji occurred. They reported that most emojis are positive, and the emojis that are more used are more emotionally loaded. Tauch et al. studied the sentiment effects on mobile phone notifications of duplicate emojis~\cite{tauch2016roles}. They discovered that when the number of emojis was high, the sentiment of the whole message was not related to the text content. According to ~\cite{miller2016blissfully}, the sentiments and interpretations of emojis may vary from people to people. The authors reported the most differently interpreted emojis in the paper.  The different usages of emojis also reveal the differences among people. It is suggested that, countries have their specific preferences to emojis~\cite{barbieri2016cosmopolitan}, and such preferences imply the cultural and regional features of the countries in many aspects~\cite{lu2016learning}. 

\subsection{Intentions of Using Emoticons \& Emojis}

In CMC, because of the lack of nonverbal cues, people introduce surrogates such as emoticons and emojis to supply these missing cues. Typical intentions of using emoticons and emojis have been discussed in the literature. Expressing emotion is one of the most studied intentions for using emoticons~\cite{lo2008nonverbal,tauch2016roles}. Also, people like to use emoticons to strengthen the expression~\cite{walther2001impacts}, and express humor~\cite{dresner2010functions}. Derkes et al.~\cite{derks2008emoticons} reported that these three intentions are the major intentions of using emoticons. In~\cite{derks2007emoticons2}, the authors discussed the functions of emoticons of expressing intimacy. Similar to face expressions, emoticons are also reported to be used to express irony~\cite{filik2015sarcasm}. In this paper, Filik et al. studied the effects of emoticon on conveying sarcasm. Another intention that emojis are used for is adjusting tone -- people appeared to use emojis to make their messages less serious and more friendly~\cite{walther2001impacts,derks2007emoticons}. The linguistic functions of emojis are discussed in~\cite{cramer2016sender,kelly2015characterising}. It suggested that emojis sometimes were used as replacements of words, and to describe contents.

\subsection{Sentiment Effects of Emoticons}
There is much work in psychology focusing on the sentiment effects of emoticons. We borrow the approaches suggested in these studies, since emojis can be considered the natural evolution of emoticons. Walther et al. studied the sentiment impacts of emoticons in CMC~\cite{walther2001impacts}. In this work, for the first time, they proposed to study emoticons and plain verbal messages as a whole. They studied the impacts of positive and negative emoticons on positive and negative verbal messages. In the paper, it is reported that positive emoticons increase the positivity of positive verbal messages, but negative emoticon do not increase the negativity of negative messages. Following the same approach, Derks et al. studied the sentiment impacts of more types of emoticons in various social contexts, and reported similar results~\cite{derks2007emoticons,derks2007emoticons2}. By applying similar approaches, the influences of emoticons on person perception~\cite{ganster2012same}, and the effects of emoticons in task-oriented communication~\cite{luor2010effect} were also studied.

\section{User Study Design}
\subsection{Emoji Selection}
Since facial emojis are the most commonly used emoji type by people~\cite{miller2016blissfully}, in this work, we study three types of facial emojis -- positive, neutral and negative -- and select five representatives for each type. Meanwhile, since the intentions and effects of using non-facial emojis are still unclear~\cite{kelly2015characterising}, we also study five representatives of this type of emojis. All the emojis studied in this paper are in IOS format -- one of the most employed emoji formats\footnote{https://emogi.com/documents/Emoji\_Report\_2015.pdf}. To select representative emojis for each type, we first rank the popularities of all emojis. The ranking is obtained according to the amount of occurrences of each emojis in our pre-collected dataset. The pre-collected dataset contains 10 million tweets, and were collected from May to September 2016 using the Twitter Streaming API\footnote{https://dev.twitter.com/streaming/overview}. To select facial emojis of different valences, we follow the approach reported in~\cite{luor2010effect}. We collect the top 50 most commonly used facial emojis. The students in a class are asked for voluntary participations to assign one of the three valences to each emoji. To avoid the divergent appearances of emojis in different platforms, all the emojis are displayed to the participants in the form of fixed pictures instead of the original Unicode graphics. 38 students participate the survey, 55.3\% of them are male, 44.7\% are female, and the average age is 23.4 ($SD$ = 2.7). We then select the top five emojis that are assigned as positive by most students as the representatives of positive type, and the same for neutral and negative types. For example, \raisebox{-0.1em}{\includegraphics[height=0.9em]{1F60A.png}} is assigned by 97\% students as positive, and is selected as one of the representatives of positive emojis. The agreement levels on the valences of most of these 15 preventatives are higher than 80\%, and all agreement levels are higher than 65\%. Table~\ref{tab:emojis} summarizes the representatives of each type, as well as the percentage of students who assign the representative emojis to the corresponding types. As to the non-facial type, we select the top five most used non-facial emojis as its representatives. Although \raisebox{-0.1em}{\includegraphics[height=0.9em]{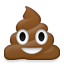}} is quite popular (ranked third), we exclude it in our study because it is not strictly non-facial. Neither do we include \raisebox{-0.1em}{\includegraphics[height=0.9em]{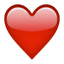}} (ranked fourth), because the functions and meanings of this emoji is well studied~\cite{lu2016learning}. Table~\ref{tab:emojis} also lists the selected representatives of non-facial emojis. 

\begin{table} 
\centering
\begin{tabular}{ | l || c | c | c | c | c | }
\hline
Emoji Type & \multicolumn{5}{c|}{ Emoji \& Percentage}\\
\hline \hline
\multirow{2}{*}{Positive} & \raisebox{-0.1em}{\includegraphics[height=0.9em]{1F60A.png}} & \raisebox{-0.1em}{\includegraphics[height=0.9em]{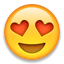}} & \raisebox{-0.1em}{\includegraphics[height=0.9em]{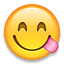}} & \raisebox{-0.1em}{\includegraphics[height=0.9em]{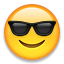}} & \raisebox{-0.1em}{\includegraphics[height=0.9em]{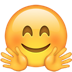}} \\
                     & 97\%                 &  97\%                &  97\%                & 94\%                 & 91\%                 \\
\hline
\multirow{2}{*}{Neutral} & \raisebox{-0.1em}{\includegraphics[height=0.9em]{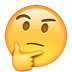}} & \raisebox{-0.1em}{\includegraphics[height=0.9em]{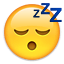}} & \raisebox{-0.1em}{\includegraphics[height=0.9em]{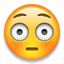}} & \raisebox{-0.1em}{\includegraphics[height=0.9em]{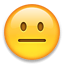}} & \raisebox{-0.1em}{\includegraphics[height=0.9em]{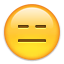}} \\
                     & 86\%                 &  80\%                &  74\%                & 66\%                 & 66\%                 \\
\hline
\multirow{2}{*}{Negative} & \raisebox{-0.1em}{\includegraphics[height=0.9em]{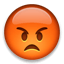}} & \raisebox{-0.1em}{\includegraphics[height=0.9em]{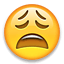}} & \raisebox{-0.1em}{\includegraphics[height=0.9em]{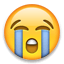}} & \raisebox{-0.1em}{\includegraphics[height=0.9em]{1F614.png}} & \raisebox{-0.1em}{\includegraphics[height=0.9em]{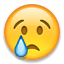}} \\
                     & 94\%                 &  94\%                &  91\%                & 83\%                 & 66\%                 \\
\hline
          Non-facial & \raisebox{-0.1em}{\includegraphics[height=0.9em]{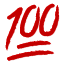}} & \raisebox{-0.1em}{\includegraphics[height=0.9em]{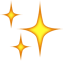}} & \raisebox{-0.1em}{\includegraphics[height=0.9em]{1F389.png}} & \raisebox{-0.1em}{\includegraphics[height=0.9em]{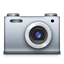}} & \raisebox{-0.1em}{\includegraphics[height=0.9em]{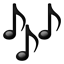}}  \\
\hline
\end{tabular}
\caption{Representative emojis of four emoji types. For the facial emojis (positive, neutral, and negative), we also list the percentages of human subjects who assign the emojis to the corresponding types.}~\label{tab:emojis}
\end{table}

\begin{table*}
\centering
\begin{tabular}{ |l | l | }
\hline
\hline
Condition & Combined Meg.\\
\hline
\hline
\textit{Positive control condition} & Just attended a seminar, it's a joy. I wish all seminars were just like it.\\
\hline
\textit{Positive meg. + a single emoji} &  Just attended a seminar, it's a joy. I wish all seminars were just like it.\raisebox{-0.1em}{\includegraphics[height=0.9em]{1F60A.png}} \\                   
\hline
\textit{Positive meg. + duplicate emojis} &  Just attended a seminar, it's a joy. I wish all seminars were just like it.\raisebox{-0.1em}{\includegraphics[height=0.9em]{1F60A.png}}\raisebox{-0.1em}{\includegraphics[height=0.9em]{1F60A.png}}\raisebox{-0.1em}{\includegraphics[height=0.9em]{1F60A.png}}\\
\hline
\hline
\textit{Neutral control condition} & Just attended a seminar, now heading back home.\\ 
\hline
\textit{Neutral meg. + a single emoji} & Just attended a seminar, now heading back home.\raisebox{-0.1em}{\includegraphics[height=0.9em]{1F60A.png}}\\
\hline
\textit{Neutral meg. + duplicate emojis} & Just attended a seminar, now heading back home.\raisebox{-0.1em}{\includegraphics[height=0.9em]{1F60A.png}}\raisebox{-0.1em}{\includegraphics[height=0.9em]{1F60A.png}}\raisebox{-0.1em}{\includegraphics[height=0.9em]{1F60A.png}}\\
\hline
\hline
\textit{Negative control condition} & Just attended a seminar, it's a hell. I wish I never have another seminar like it.\\
\hline
\textit{Negative meg. + a single emoji} & Just attended a seminar, it's a hell. I wish I never have another seminar like it.\raisebox{-0.1em}{\includegraphics[height=0.9em]{1F60A.png}}\\
\hline
\textit{Negative meg. + duplicate emojis} & Just attended a seminar, it's a hell. I wish I never have another seminar like it.\raisebox{-0.1em}{\includegraphics[height=0.9em]{1F60A.png}}\raisebox{-0.1em}{\includegraphics[height=0.9em]{1F60A.png}}\raisebox{-0.1em}{\includegraphics[height=0.9em]{1F60A.png}}\\
\hline
\end{tabular}
\caption{Examples of combined messages displayed to the participants in the user study for positive emojis. We only list the combinations of plain verbal messages and emoji \raisebox{-0.1em}{\includegraphics[height=0.9em]{1F60A.png}} in the table. The plain verb messages are borrowed from~\cite{walther2001impacts}, with minor adjustments to fit our study.}~\label{tab:examples}
\end{table*}

\subsection{Intention Selection}
By reviewing the previous work on the intentions of using nonverbal cues as their surrogates in communication, seven intentions are summarized. We summarize these intentions and their explanations from previous work as follows.

\begin{itemize}
\item \textit{\textbf{Expressing sentiment}}: similar to facial expressions in face-to-face communication, using emojis in CMC to express sentiments or emotions, such as anger, happiness, fear, and so on.

\item \textit{\textbf{Strengthening expression}}: using emojis to strengthen the expression, for example, making a positive post or message more positive by adding emojis.

\item \textit{\textbf{Adjusting tone}}: using emojis to adjust tone, for example, making a post or message more friendly or less serious by adding emojis.

\item \textit{\textbf{Expressing humor}}: using emojis to make communication more funny and lively.

\item \textit{\textbf{Expressing irony}}: using emojis to make communication more sarcastic or ironic.

\item \textit{\textbf{Expressing intimacy}}: using emojis to make the readers of the posts or messages feel more engaged, express intimacy and closeness to the readers.

\item \textit{\textbf{Describing content}}: using emojis to describe content in text, for example, using a national flag emoji to denote a country or related content.

\end{itemize}

\subsection{User Study on the Intentions of Using Emojis}

Following the approach of studying message senders' intentions of using emoticons reported in~\cite{derks2008emoticons}, we construct a 7 (intentions)$\times$20 (emojis) within-subjects online survey in this study. For each combination of an emoji and an intention, we ask the participants to rate their willingness to use the emoji for the intention on a 7-point scale (7 = totally willing, 1 = not willing at all). For each participant, the 140 combinations of emojis and intentions in a survey are displayed in a random order. Please note that the emojis are displayed as pictures instead of Unicode graphics to avoid the influences of different displaying platforms. We also ask the participants to report their basic demographic information, and frequencies of using emojis. We post the link to the online survey on AMT (Amazon Mechanic Turk), and require participants to be \textit{Master} workers to ensure the quality of the results. The estimated time for completing the survey is 13 minutes, and we compensate each participant with \$2. Seventy five workers participated the study. We check the qualities of the responses, and manually filter out the problematic responses, for example, when the answers to all the questions are the same. After the quality check, we retained the data of 69 participants. 63.8\% of them are male, and the remaining 36.2\% are female. The average age of participants is 36.5 ($SD$ = 7.7). Among them, 21.7\% workers report that they rarely or never use emojis, 36.2\% workers use emojis most of time or always, and the remaining use emojis sometimes. We believe this is a reasonable mix of different levels of experience with emojis. Note that emojis are meant to be intuitive even to the inexperienced users.

\subsection{User Study on the Sentiment Effects of Emojis}
To study the sentiment effects of emojis on message recipients, we construct four separate user studies. Each user study corresponds to a type of emojis. As suggested in previous work~\cite{derks2007emoticons,ganster2012same}, we prepare three plain verbal messages of positive, negative, and neutral valences. The three messages are borrowed from~\cite{walther2001impacts} with minor adjustments to fit our study. In our surveys, a plain message is combined with either an emoji, or three duplicate emojis, or as a control condition with no emoji. By analyzing the pre-collected data, we find that people more often use an emoji three times in a row than other number of duplications, and the average number of duplications is 3.3. Accordingly, we set the number of duplications of an emoji as three in our survey. Therefore, each user study is constructed as a 3 (messages: positive, neutral, negative) $\times$ 11 (emojis of a type: five single emojis, five duplicate emojis, no emoji) within-subjects survey, resulting in 33 questions. Table~\ref{tab:examples} shows some example questions in the survey of positive emojis. For each question, we ask the participants to rate the sentiment of the combination of plain message and emoji(s) on a 7-points scale (7 = very positive, 1 = very negative). These questions are also displayed in random orders to participants. We post the links to four surveys on AMT, and ask for participations of different master workers. The estimated time for completing a survey is 6 minutes, and we compensate each participant with \$1. For each survey, we collect 75 responses. After quality check, we filter out 13 problematic responses, and retain 287 responses (70, 72, 73, and 73 responses for each survey, respectively). 54.7\% of the 287 participants are male, and the remaining 45.3\% are female. The average age of the participants are 37.3 ($SD$ = 8.5). Twenty three percent of them report that they rarely or never use emojis in communication, 31.3\% use emojis most of time or always use emojis, and the remaining use emojis sometimes.

\begin{table*}

\centering
\begin{tabular}{ | c || c | c | c | c | c | }
\hline
Intention & Top 5 Proper Emojis & Ranking of Emoji Types\\
\hline \hline
Expressing Sentiment & \raisebox{-0.1em}{\includegraphics[height=0.9em]{1F621.png}}, \raisebox{-0.1em}{\includegraphics[height=0.9em]{1F614.png}}, \raisebox{-0.1em}{\includegraphics[height=0.9em]{1F622.png}}, \raisebox{-0.1em}{\includegraphics[height=0.9em]{1F60A.png}}, \raisebox{-0.1em}{\includegraphics[height=0.9em]{1F62D.png}} & Negative $>$ Positive $>$ Neutral $>$ Non-facial\\
\hline
Strengthening the Expressions & \raisebox{-0.1em}{\includegraphics[height=0.9em]{1F621.png}}, \raisebox{-0.1em}{\includegraphics[height=0.9em]{1F60A.png}}, \raisebox{-0.1em}{\includegraphics[height=0.9em]{1F622.png}}, \raisebox{-0.1em}{\includegraphics[height=0.9em]{1F633.png}}, \raisebox{-0.1em}{\includegraphics[height=0.9em]{1F62D.png}} & Positive = Negative $>$ Neutral $>$ Non-facial\\
\hline
Adjusting Tone  & \raisebox{-0.1em}{\includegraphics[height=0.9em]{1F60A.png}}, \raisebox{-0.1em}{\includegraphics[height=0.9em]{1F621.png}}, \raisebox{-0.1em}{\includegraphics[height=0.9em]{1F60E.png}}, \raisebox{-0.1em}{\includegraphics[height=0.9em]{1F60B.png}}, \raisebox{-0.1em}{\includegraphics[height=0.9em]{1F622.png}} & Positive = Negative $>$ Neutral $>$ Non-facial\\
\hline
Describing Content & \raisebox{-0.1em}{\includegraphics[height=0.9em]{1F389.png}}, \raisebox{-0.1em}{\includegraphics[height=0.9em]{1F634.png}}, \raisebox{-0.1em}{\includegraphics[height=0.9em]{1F622.png}}, \raisebox{-0.1em}{\includegraphics[height=0.9em]{1F60A.png}}, \raisebox{-0.1em}{\includegraphics[height=0.9em]{1F4F7.png}} & Non-facial $>^*$ Neagtive = Positive = Neutral\\
\hline
Expressing Humor & \raisebox{-0.1em}{\includegraphics[height=0.9em]{1F60B.png}}, \raisebox{-0.1em}{\includegraphics[height=0.9em]{1F60A.png}}, \raisebox{-0.1em}{\includegraphics[height=0.9em]{1F917.png}}, \raisebox{-0.1em}{\includegraphics[height=0.9em]{1F60E.png}}, \raisebox{-0.1em}{\includegraphics[height=0.9em]{1F633.png}} & Positive $>$ Neutral $>$ Negative $>$ Non-facial\\
\hline
Expressing Intamacy & \raisebox{-0.1em}{\includegraphics[height=0.9em]{1F60D.png}}, \raisebox{-0.1em}{\includegraphics[height=0.9em]{1F917.png}}, \raisebox{-0.1em}{\includegraphics[height=0.9em]{1F60A.png}}, \raisebox{-0.1em}{\includegraphics[height=0.9em]{1F60B.png}}, \raisebox{-0.1em}{\includegraphics[height=0.9em]{1F622.png}} & Positive $>$ Negative = Neutral = Non-facial\\
\hline
Expressing Irony & \raisebox{-0.1em}{\includegraphics[height=0.9em]{1F914.png}}, \raisebox{-0.1em}{\includegraphics[height=0.9em]{1F610.png}}, \raisebox{-0.1em}{\includegraphics[height=0.9em]{1F611.png}}, \raisebox{-0.1em}{\includegraphics[height=0.9em]{1F633.png}}, \raisebox{-0.1em}{\includegraphics[height=0.9em]{1F60E.png}} & Neutral $>$ Positive $>$ Negative $>$ Non-facial\\
\hline
\end{tabular}

\caption{Seven intentions ranked from top to bottom according to their popularities, in terms of using emojis for. The table also list the top five most proper emojis, as well as the group mean scores ranking of four emojis types for each intention. $^*$All the inequalities reported in the table are significant at the 0.05 level, except for the noted one, which is marginally significant (p-value = 0.052).}~\label{tab:intentions}
\end{table*}

\section{Preliminary}
In this section, we briefly introduce the statistical backgrounds involved in this paper. 

\subsubsection{Within- \& Between-subjects Design}
In a within-subjects experimental design, participants are asked to take all treatments (e.g. emojis in our user studies). For example, in our first user study of user intention, workers are asked to rate all 20 emojis for all seven intentions. Therefore, this user study is a within-subjects study. On the contrary, in a between-subjects experimental design, each participant is only asked to take one treatment, for example, asking a worker rate only one emoji. Clearly, within-subjects experimental design could effectively reduce the amount of participants. However, this experimental design also introduces dependency to the collected data -- the 7 $\times$ 20 scores rated by one worker are not independent. To deal with the dependency, a repeated measure ANOVA, also referred to as a within-subjects ANOVA, is suggested to be used. 
Similar to a normal ANOVA, a repeated measure ANOVA is used to analyze the differences among group means, but takes the dependency in data into account.

\subsubsection{Tukey Post Hoc Tests}
An ANOVA test analyzes the differences among group means. However, it only provides information about if the groups means are equal (accept null hypothesis), or different (reject null hypothesis). When the null hypothesis is rejected, to further analyzes the differences between groups, a Tukey post hoc test or other types of post hoc tests is usually conducted. 
A Tukey post hoc test compares all the group means in pairs, and report the significances of the pairwise differences. Comparing with pairwise comparisons of group means using t-tests, a Tukey post hoc test avoids the accumulation of Type I error. For example, in each survey of our second user study of emoji sentiment effects, a repeated measure ANOVA indicates the significant differences between the combinations of emojis and plain verbal messages. Therefore, we conducted Tukey post hoc tests to analyze these combinations in pairs to study emoji sentiment in detail. 

\subsubsection{Statistical Contrast Tests}
A contrast test is used to compare the difference between combinations of treatments. 
It is also a type of post hoc test, and usually applied when a ANOVA test is conducted and rejects the null hypothesis (i.e. significant differences exist among treatments). In a linear contrast test, treatments are linearly combined according to prior knowledge, for example, averaging certain treatments. A linear combination of treatments are referred as a contrast. The result of a contrast test indicates if the difference between two combinations of treatments is significant. Since we have obtained the prior knowledge of the valences of 20 emojis, we are able to combine them into four groups. In our user study of user intentions, we conduct linear contrast tests to study the differences of willingness scores for each intention between emoji types.

\section{Intentions of Using Emoji}
We first study the intentions of using emojis from a perspective of message senders. In the first user study, workers are asked to rate the willingness scores of using an emoji for an intention. We use the mean value of the willing scores to measure how proper the emojis are for the intentions. For example, if the mean score of an emoji for an intention is significantly\footnote{Unless otherwise specified, the significant level is defined at the 0.05 level in this paper.} higher than 6 points (out of 7), then people are very willing to use the emoji to achieve the intention. 4 points, the median of the rating scale, is taken as the threshold for judging if an emoji is considered proper or not for an intention. The popularity of an intention in terms of using emojis for, is then measured as its number of proper emojis. The more emojis that are considered proper, the more popular the intention is. If two intentions have same number of proper emojis, we compare their numbers of very proper emojis, of which mean scores are significantly higher than 6 points. The comparisons between the mean values of willingness scores with constants are conducted using one sample one-tailed t-tests.

We also study the differences between different types of emojis (positive, neutral, negative, and non-facial). A one-way repeated measure ANOVA is first performed for each intention, to test the equivalence between the mean values of the scores of all the emojis. The tests on all intentions report significant differences between these mean values, indicating that the emojis are used with divergent willingness levels for the intentions. Therefore, we construct pairwise linear contract tests between emoji types for each intention. A contrast is computed as the average of the scores of the five emojis of a type. For an intention, if a contrast test indicates that the mean score of a type of emoji is significantly higher than that of another type, then the former type is considered more proper. If no significant difference is detected, then the two types are considered equally proper/improper for the intention. Please note that, for the conciseness of the narration in this paper, we do not list the p-values of all the statistical tests. The p-values of the significant differences reported in the paper are all lower than 0.05. We summarize the results in Table~\ref{tab:intentions}. Means and standard deviations of sentiment scores of the combinations of emojis and intentions are listed in appendix section Table~\ref{tab:allemojis}.


\subsubsection{Expressing Sentiment}
As the designers' original intent of introducing emojis~\cite{dix2007designing}, expressing sentiment is the most popular intention for using emojis. The mean scores of 15 emojis are significantly higher than 4 points, indicating that a majority of emojis are proper for conveying sentiment. There is only one non-facial emoji (\raisebox{-0.1em}{\includegraphics[height=0.9em]{1F389.png}}) rated higher than 4 points, implying that this type of emoji are not often used for this intention. Meanwhile, seven emojis' scores are significantly higher than 6 points, revealing that people are very willing to use them to express sentiment. It is observed that, for this intention, negative emojis are more preferred by message senders -- all five negative emojis are listed among the most rated seven emojis, and the rest two are both positive emojis. Contrast tests confirm our observations. Mean score of negative emojis ($M$ = 6.57, $SD$ = 0.59) is significantly higher than those of other groups of emojis. Therefore, it is suggested that people are indeed more intended to use negative emojis to express sentiment. Among the remaining three types, positive emojis ($M$ = 6.09, $SD$ =  0.75) are rated higher than neutral emojis ($M$ = 4.82, $SD$ =  1.17), and non-facial emojis ($M$ = 2.59, $SD$ =  0.83) are the least.

\subsubsection{Strengthening Expression}
Strengthening expression is the second popular intention, with 14 emojis' mean willingness scores rated significantly higher than 4 points, but only one higher than 6 points (\raisebox{-0.1em}{\includegraphics[height=0.9em]{1F621.png}}). Non-facial emojis are still not often used for this intention -- \raisebox{-0.1em}{\includegraphics[height=0.9em]{1F389.png}} is again the only non-facial emoji rated high than 4 points. According to the results of contrast tests, both positive ($M$ = 5.56, $SD$ =  0.94) and negative emojis ($M$ = 5.61, $SD$ =  1.26) are scored significantly higher than neutral emojis ($M$ = 4.42, $SD$ = 1.37), followed by the non-facial type ($M$ = 3.23, $SD$ = 1.02). However, the difference between positive and negative emojis is not significant (p-value = 0.65), indicating that people are equally willing to use these two types to strengthen the expression in communication. 

\subsubsection{Adjusting Tone}
The same 14 emojis that are rated higher than 4 points for strengthening expression are rated higher than 4 points for adjusting tone. However, there is no emoji scored significantly higher than 6 points, indicating the lower popularity of this intention. In terms of types of emojis, we discover that positive ($M$ = 5.14, $SD$ = 1.19) and negative emojis ($M$ = 4.98, $SD$ = 1.33) are equally intended to be used for adjusting tone, and both of them are significantly higher rated than neutral emojis ($M$ = 4.42, $SD$ = 1.31). Non-facial emojis ($M$ = 2.8, $SD$ = 0.91) are again the least scored. 

\subsubsection{Describing Content}
Describing content is the fourth popular intention. Only five emojis' mean willingness scores are significantly higher than 4 points, and no emoji's score exceeds 6 points. The small number of proper emojis and their relatively low mean scores indicate that people usually do not use emojis to achieve this intention, or the other intentions that are even less popular. Aligned with the conclusion in~\cite{tauch2016roles}, the mean scores of non-facial ($M$ = 4.31, $SD$ = 1.35) emojis is highest among four types, indicating that this emoji type is the most proper to describe content in communications. It is significantly higher than the mean scores of positive ($M$ = 3.87, $SD$ = 1.6) and neutral emojis ($M$ = 3.85 , $SD$ = 1.4), and marginally significantly higher (p-value = 0.052) than that of negative emojis ($M$ = 4.03, $SD$ = 1.68). Meanwhile, there is no significant difference discovered between positive, negative, and neutral emojis in terms of mean scores.

\subsubsection{Expressing Humor}

Although expressing humor is one of the most popular intentions of using emoticon~\cite{derks2008emoticons}, people are not that intended to use emojis for this intention. Only four emojis' scores exceed 4 points, and no one of them has a mean willingness score significantly higher than 6 points. It is observed that all these four emojis are positive emojis. Contrast tests suggest that the mean score of positive emojis ($M$ = 4.56, $SD$ = 0.15) is indeed significantly higher than other groups, indicating that this type is the most proper to express humor. Among the remaining three types, neutral type ($M$ = 3.09, $SD$ = 1.40) is rated higher than negative type ($M$ = 2.12, $SD$ = 1.10), and both of them are significantly higher rated than non-facial type ($M$ = 1.91, $SD$ = 0.85).

\subsubsection{Expressing Intimacy}
Among all the emojis, only \raisebox{-0.1em}{\includegraphics[height=0.9em]{1F60D.png}} is rated significantly higher than 4 points ($M$ = 5.94, $SD$ = 1.38). The uniqueness of this high willingness score reveals that people are particularly willing to use this emoji to express intimacy. We suggest that this is because of the \raisebox{-0.1em}{\includegraphics[height=0.9em]{2764.png}} elements in the emoji. Contrast tests suggest that the mean score of positive emojis ($M$ = 3.56, $SD$ = 1.05) is significantly higher than other groups. There is no significant difference found between neutral ($M$ = 1.62, $SD$ = 0.90), negative ($M$ = 1.97, $SD$ = 1.05), and non-facial ($M$ = 1.44, $SD$ = 0.66) emojis.

\subsubsection{Expressing Irony}
Expressing irony is the least popular intention among all, there are no emojis being rated a score higher than 4 points. It is worth noting that four emojis' scores are significantly higher than 3 points, and these four emojis are all neutral emojis. This observation indicates that if people have to use emojis to express irony, neutral emojis are more likely to be used. In terms of emoji types, we discover that the mean score of neutral type is the highest ($M$ = 3.38, $SD$ = 1.20). Among the remaining three, positive type ($M$ = 2.48, $SD$ = 1.24) is significantly higher rated than negative type ($M$ = 2.23, $SD$ = 1.09), and non-facial ($M$ = 1.66, $SD$ = 0.80) is the least rated.

\section{Sentiment Effects of Emojis}
In our second user study of emoji sentiment effects, four separate groups of workers participate the surveys for four emoji types. Repeated measure ANOVA tests are first conducted on the collected data. All the tests suggest significant differences between the mean sentiment scores of the combinations of emojis and plain verbal messages. To study the effect of an emoji on a message, we perform Tukey post hoc tests to compare the mean sentiment scores between two conditions: (1) the message coupled with the emoji and (2) the control condition (the same message with no emoji). A significant positive difference indicates that the emoji increases the sentiment, and a significant negative difference indicates that the emoji decreases the sentiment. Similarly, to study the sentiment effects of duplicate emojis, we compare the mean sentiment scores between two conditions: (1) a single emoji coupled with a message, and (2) the duplications of this emoji coupled with the same message. 

We pay particular attention to two types of combinations: plain positive messages with negative emojis, and plain negative messages with positive emojis. For these combinations, we compare their mean sentiment scores with 4 points (neutral sentiment), to study which one determines the sentiment of the combined message, emojis or plain verbal messages. The comparisons are performed using one sample one-tailed t-tests. Take combination of the plain positive message and a negative emoji for example. If its mean sentiment score is significantly higher than 4 points, then the overall sentiment of the combined message is positive, indicating the plain message contributes more, in terms of sentiment. If the score is significantly lower than 4 points, it is the emoji that determines the overall sentiment. If there is no significant difference between the mean score and 4 points, then the message and emoji neutralize each other. 

It is also worth noting that the mean sentiment scores of the plain positive, neutral, and negative verbal messages are (6.26, 4.20, 1.81), (6.13, 4.15, 1.75), (6.11, 4.00, 1.55), and (6.44, 4.25, 1.38) in four surveys\footnote{The corresponding standard deviations of them are (1.14, 0.69, 1.25), (1.05, 0.56, 1.22), (0.89, 0.71, 0.86), and (1.10, 0.76, 0.65), respectively.}. This indicates the consistency of workers' scales cross surveys, as well as the effectiveness of the control conditions. We discuss increments or decrements of sentiment scores brought by emojis in this section. We summarize the results in Table~\ref{tab:sentiment}. The mean values and standard deviations of sentiment scores of four surveys are reported in appendix section Table~\ref{tab:positiveemojis} -~\ref{tab:non-facialemojis}. 

\subsection{Positive Emojis}

\subsubsection{Positive Message}
The results of Tukey post hoc tests indicate that there are no significant difference between the plain positive verbal message and the combinations of the message with five positive emojis, in terms of mean sentiment score. Such results suggest that positive emojis coupled with a positive verbal message do not convey greater positivity. It also implies the difference between positive emojis and positive emoticons. The later are reported having increasing effects on positive verbal messages~\cite{walther2001impacts}.

\subsubsection{Neutral Message}

All five positive emojis increase the sentiment score of the plain neutral message. The largest increment is brought by \raisebox{-0.1em}{\includegraphics[height=0.9em]{1F60D.png}} , with a mean increment of 2.04 ($SD$ = 0.14), and the smallest is brought by \raisebox{-0.1em}{\includegraphics[height=0.9em]{1F60E.png}} ($M$ = 1.27, $SD$ = 0.12).

\subsubsection{Negative Message}

Positive emojis significantly increase the sentiment of the plain negative message. Again, \raisebox{-0.1em}{\includegraphics[height=0.9em]{1F60D.png}} is the most effective emojis for increasing sentiment ($M$ = 1.61, $SD$ = 0.17), and \raisebox{-0.1em}{\includegraphics[height=0.9em]{1F60E.png}} is the least ($M$ = 1.19, $SD$ = 0.14). We also conduct five one-sample one-tailed t-tests to compare the sentiment scores of these combinations with 4 points. The results indicate that the sentiment of them are all significantly lower than 4 points. Therefore, although positive emojis coupled with negative verbal messages convey less negativity, the overall sentiments are determined by the verbal part, and are still negative.

\subsection{Neutral Emojis}
\subsubsection{Positive Message}
Interestingly, neutral emojis have significant decreasing effects on the sentiment of the plain positive message. Emoji \raisebox{-0.1em}{\includegraphics[height=0.9em]{1F634.png}} brings the largest decrement ($M$ = -2.11, $SD$ = 0.20), and \raisebox{-0.1em}{\includegraphics[height=0.9em]{1F633.png}} brings the least ($M$ = -1.19, $SD$ = 0.19). According to our study in intentions, neutral emojis are the most proper to express irony. We suppose that neutral emojis may introduce sarcasm to a plain positive message, and consequently, decrease the positivity.

\subsubsection{Neutral Message}
It is suggested that neutral emojis do not have significant effects on a neutral message. All five Tukey post hoc tests report no significant difference between a plain neutral message and combinations of neutral emojis and the message, in terms of mean sentiment score. 

\subsubsection{Negative Message}
Similar to the neutral message, no significant effect is found of neutral emojis on a negative message. Therefore, the decreasing effects of neutral emojis are only observed on a plain positive message. 

\begin{table}
\centering

\begin{tabular}{|c||c|c|c|}
\hline
                          &Meg$_{pos}$ &Meg$_{neu}$ &Meg$_{neg}$ \\
\hline
\hline
Emojis$_{pos}$ & No effect & Increasing & Increasing\\
\hline
Emojis$_{neu}$ & Decreasing & No effect & No effect\\
\hline
Emojis$_{neg}$ & Decreasing$^*$ & Decreasing & No effect\\
\hline
\end{tabular}


\caption{The effects of three types of facial emojis on three valences of plain verbal messages. We do not include non-facial emojis in this table, since their sentiment effects on plain messages are not consistent. $^*$The overall sentiment of the combination of negative emojis and the plain positive verbal message are decreased to neutral or negative.}~\label{tab:sentiment}


\end{table}

\subsection{Negative Emojis}
\subsubsection{Positive Message}

All five negative emojis significantly decrease the sentiment of a plain positive message, with the largest decrement of \raisebox{-0.1em}{\includegraphics[height=0.9em]{1F621.png}} ($M$ = -3.18, $SD$ = 0.24), and smallest decrement of \raisebox{-0.1em}{\includegraphics[height=0.9em]{1F622.png}} ($M$ = -2.32, $SD$ = 0.20). We also compute the overall valences of these combinations. The results show that the sentiment of the combination of \raisebox{-0.1em}{\includegraphics[height=0.9em]{1F621.png}} and the positive message is significantly lower than 4 points, indicating negative valence. The sentiment scores of other four combinations have no significant difference from 4 points, indicating neutral valence. The results suggest that negative emojis can either neutralize the sentiment of a positive message resulting in a neutral valence, or turn the overall sentiment to negative. Therefore, negative emojis are more powerful than positive emojis in expressing sentiment, which is consistent with our result in the earlier intention study -- people are more willing to use negative emojis to express sentiment than positive emojis.

\subsubsection{Neutral Message}
Tukey post hoc tests report that negative emojis also have significant decreasing effects on plain neutral messages, but not as great as on plain positive messages. The largest decrement is brought by \raisebox{-0.1em}{\includegraphics[height=0.9em]{1F621.png}} ($M$ = -1.97, $SD$ = 0.14), and the smallest by \raisebox{-0.1em}{\includegraphics[height=0.9em]{1F614.png}} ($M$ = -0.38, $SD$ = 0.14).

\subsubsection{Negative Message}
Similar to positive emojis on a plain positive message, negative emojis also do not have sentiment effects on a plain negative message. The tests results show that the sentiment scores of these combinations are not significantly different from that of a plain message.

\subsection{Non-facial Emojis}
Our results of positive, neutral, and negative emojis suggest that the sentiment effects of emojis of one type are consistent. However, the sentiment effects on plain messages of non-facial emojis vary from emojis to emojis. Meanwhile, it is also observed that the effects of non-facial emojis, if any, are relatively small (lower than 1 point).

\subsubsection{Positive Message}
Test results show that two non-facial emojis have increasing sentiment effects on a plain positive message, while increment are both relatively small. These two are \raisebox{-0.1em}{\includegraphics[height=0.9em]{1F3B6.png}} ($M$ = 0.47, $SD$ = 0.10), and \raisebox{-0.1em}{\includegraphics[height=0.9em]{2728.png}} ($M$ = 0.49, $SD$ = 0.08). The remaining three emojis are reported having no significant sentiment effects on a plain message.

\subsubsection{Neutral Message}
Except for \raisebox{-0.1em}{\includegraphics[height=0.9em]{1F4F7.png}}, all non-facial emojis are reported significantly increasing the sentiment of a plain neutral message. Again, the increments are relatively small, with the largest brought by \raisebox{-0.1em}{\includegraphics[height=0.9em]{1F389.png}} ($M$ = 0.93, $SD$ = 0.11), and smallest brought by \raisebox{-0.1em}{\includegraphics[height=0.9em]{2728.png}} ($M$ = 0.71, $SD$ = 0.09). Emoji \raisebox{-0.1em}{\includegraphics[height=0.9em]{1F4F7.png}} does not significantly affect the sentiment in this case.

\subsubsection{Negative Message}
All five Tukey post hoc tests report that adding non-facial emojis to the plain negative message does not result in significant differences. Therefore, non-facial emojis do not affect the sentiment of a negative message.

\subsection{Duplicate Emojis}

Different from the conclusion reported in~\cite{tauch2016roles}, our results suggest that, in most cases, duplicating do not make significant differences over single emojis, in terms of sentiment score. Under all circumstances, the usage of duplicate positive and neutral emojis do not increase or decrease sentiment scores. For negative emojis, we observe that the duplications of two emojis convey more negativity, but only when coupled with the plain negative messages. The decreasing effects are relatively small, and these two negative emojis are  \raisebox{-0.1em}{\includegraphics[height=0.9em]{1F629.png}} ($M$ = -0.39, $SD$ = 0.09), and \raisebox{-0.1em}{\includegraphics[height=0.9em]{1F614.png}} ($M$ = -0.32, $SD$ = 0.06). The duplications of non-facial are reported expressing more positivity than single emoji only in two cases, both coupled with the plain positive message. The duplication of \raisebox{-0.1em}{\includegraphics[height=0.9em]{1F4AF.png}} are more positive than its single version ($M$ = 0.45 , $SD$ = 0.08), so does the duplication of \raisebox{-0.1em}{\includegraphics[height=0.9em]{1F389.png}} ($M$ = 0.31, $SD$ = 0.06). It is suggested that duplications of emojis only enhance the sentiment under limited circumstances, and the enhancements are relatively small. 

\section{Conclusion}
In this work, we extensively study the usage of emojis in CMC from both the perspectives of message senders and recipients.  We construct two user studies, and extensive statistical hypothesis tests are conducted to analyze the collected data. From the perspective of senders, we focus on the intentions of using emojis. We discover that the most popular intentions are expressing sentiment, strengthening expression, and adjusting tone. Moreover, our result suggest the subtle differences between emoji types, such as negative emojis are more intended to be used than positive emojis to express sentiment, and neutral emojis are the most proper to express irony. From a message recipient perspective, we study how recipients feel about the sentiment conveyed by emojis. In stead of separating emojis and verbal messages, we treat these two parts as a whole.  Our results uncover the divergent sentiment effects of emojis: positive emojis do not affect the sentiment of plain positive verb messages, nor do negative emojis affect the sentiment of plain negative messages. In addition, the results suggest that the duplicate usage of emojis do not express more intense sentiment than single emoji in most cases. Our work provide valuable insights into the usage of emojis, and shed better light on this increasingly popular type of nonverbal cue surrogates in communication.


\bibliographystyle{aaai}

\bibliography{reference}


\begin{appendices}

\begin{table*} [!htbp]
\begin{tabular}{|c|c|c|c|c|c|c|c|c|c|c|c|c|c|c|c|}
\hline
  &  &  \multicolumn{2}{c|}{Intention1} & \multicolumn{2}{c|}{Intention2} & \multicolumn{2}{c|}{Intention3} &  \multicolumn{2}{c|}{Intention4} &  \multicolumn{2}{c|}{Intention5} &  \multicolumn{2}{c|}{Intention6} &  \multicolumn{2}{c|}{Intention7} \\
\hline
 & Emoji& $M$ & $SD$ & $M$ & $SD$ & $M$ & $SD$ & $M$ & $SD$ & $M$ & $SD$ & $M$ & $SD$ & $M$ & $SD$\\
\hline
\multirow{5}{*}{Pos} & \raisebox{-0.1em}{\includegraphics[height=0.9em]{1F60A.png}} & 6.64 & 0.69 & 5.84 & 1.30 & 5.51 & 1.39 & 4.62 & 1.73 & 2.43 & 1.54 & 1.89& 3.48 & 4.30& 1.95  \\
& \raisebox{-0.1em}{\includegraphics[height=0.9em]{1F60D.png}} & 6.46 & 0.85 & 6.00 & 1.32 & 4.99 & 1.80 & 3.41 & 1.84 & 2.06 & 1.59 & 1.86& 5.94 & 4.10& 1.38  \\
& \raisebox{-0.1em}{\includegraphics[height=0.9em]{1F60B.png}} & 6.00 & 1.25 & 5.51 & 1.45 & 5.19 & 1.64 & 5.84 & 1.55 & 3.03 & 1.76 & 1.95& 2.90 & 3.84& 1.65  \\
& \raisebox{-0.1em}{\includegraphics[height=0.9em]{1F60E.png}} & 5.22 & 1.42 & 5.07 & 1.47 & 5.28 & 1.44 & 4.46 & 1.67 & 2.83 & 1.71 & 1.78& 1.90 & 3.61& 1.20   \\
& \raisebox{-0.1em}{\includegraphics[height=0.9em]{1F917.png}} & 6.13 & 1.01 & 5.38 & 1.65 & 4.75 & 1.77 & 4.48 & 1.55 & 2.06 & 1.38 & 1.90& 3.57 & 3.49 & 2.03  \\
\hline

\hline
\multirow{5}{*}{Neu} & \raisebox{-0.1em}{\includegraphics[height=0.9em]{1F914.png}} & 4.43 & 1.68 & 4.49 & 1.84 & 4.64 & 1.67 & 2.97 & 1.72 & 4.12 & 1.85 & 1.73& 1.61 & 3.90& 0.84   \\
& \raisebox{-0.1em}{\includegraphics[height=0.9em]{1F634.png}} & 4.20 & 1.88 & 4.10 & 1.96 & 4.25 & 1.83 & 3.46 & 1.98 & 2.54 & 1.73 & 2.01& 1.42 & 4.64& 0.85   \\
& \raisebox{-0.1em}{\includegraphics[height=0.9em]{1F633.png}} & 6.12 & 1.06 & 5.68 & 1.40 & 4.67 & 1.69 & 3.71 & 2.03 & 3.30 & 1.77 & 1.90& 1.99 & 3.90& 1.43   \\
& \raisebox{-0.1em}{\includegraphics[height=0.9em]{1F610.png}} & 4.93 & 1.70 & 3.96 & 2.00 & 4.38 & 1.84 & 2.61 & 1.69 & 3.52 & 1.69 & 1.87& 1.51 & 3.59& 0.90   \\
& \raisebox{-0.1em}{\includegraphics[height=0.9em]{1F611.png}} & 4.43 & 1.87 & 3.88 & 1.96 & 4.17 & 2.12 & 2.70 & 1.80 & 3.42 & 1.90 & 1.79& 1.58 & 3.23& 0.93   \\
 \hline
 
\hline
\multirow{5}{*}{Neg} & \raisebox{-0.1em}{\includegraphics[height=0.9em]{1F621.png}} & 6.78 & 0.48 & 6.20 & 1.07 & 5.42 & 1.63 & 1.80 & 1.51 & 2.13 & 1.52 & 2.02& 1.42 & 4.12& 0.99   \\
& \raisebox{-0.1em}{\includegraphics[height=0.9em]{1F629.png}} & 6.36 & 0.94 & 5.46 & 1.71 & 5.00 & 1.64 & 2.26 & 1.49 & 2.25 & 1.47 & 1.95& 1.90 & 3.74& 1.25   \\
& \raisebox{-0.1em}{\includegraphics[height=0.9em]{1F62D.png}} & 6.45 & 0.92 & 5.59 & 1.67 & 4.83 & 1.85 & 3.22 & 2.21 & 2.71 & 1.96 & 1.97& 1.93 & 3.99& 1.23   \\
& \raisebox{-0.1em}{\includegraphics[height=0.9em]{1F614.png}} & 6.65 & 0.61 & 5.23 & 1.75 & 4.68 & 1.74 & 1.78 & 1.40 & 2.22 & 1.49 & 1.91& 2.01 & 3.97& 1.36   \\
& \raisebox{-0.1em}{\includegraphics[height=0.9em]{1F622.png}} & 6.61 & 0.83 & 5.55 & 1.55 & 4.99 & 1.75 & 1.57 & 1.08 & 1.84 & 1.29 & 1.94& 2.58 & 4.36& 1.76   \\
 \hline
 
\hline
\multirow{5}{*}{Non-} & \raisebox{-0.1em}{\includegraphics[height=0.9em]{1F4AF.png}} & 3.23 & 2.02 & 4.62 & 2.04 & 3.03 & 1.80 & 2.51 & 1.84 & 2.07 & 1.52 & 1.79& 1.45 & 3.99& 1.08   \\
& \raisebox{-0.1em}{\includegraphics[height=0.9em]{2728.png}} & 1.87 & 1.45 & 2.67 & 1.76 & 2.29 & 1.51 & 1.39 & 1.10 & 1.33 & 0.95 & 2.05& 1.67 & 3.43& 1.30   \\
& \raisebox{-0.1em}{\includegraphics[height=0.9em]{1F389.png}} & 4.81 & 1.74 & 4.96 & 1.72 & 4.42 & 1.87 & 2.65 & 1.70 & 2.10 & 1.53 & 1.70& 1.55 & 4.67& 1.16   \\
& \raisebox{-0.1em}{\includegraphics[height=0.9em]{1F4F7.png}} & 1.13 & 0.42 & 1.46 & 1.08 & 1.55 & 0.99 & 1.49 & 1.04 & 1.39 & 0.81 & 2.12& 1.20 & 4.29& 0.61   \\
& \raisebox{-0.1em}{\includegraphics[height=0.9em]{1F3B6.png}} & 1.93 & 1.34 & 2.45 & 1.60 & 2.72 & 1.58 & 1.51 & 0.96 & 1.42 & 0.79 & 1.94& 1.36 & 5.17& 0.94   \\
\hline
\end{tabular}
\caption{Means and standard deviations of willingness scores of each emoji for each intention. Seven intentions from left to right are: \textit{\textbf{expressing sentiment}}, \textit{\textbf{strengthening expression}}, \textit{\textbf{adjusting tone}}, \textit{\textbf{describing content}}, \textit{\textbf{expressing humor}}, \textit{\textbf{expressing intimacy}}, and \textit{\textbf{expressing irony}}.}~\label{tab:allemojis}
\end{table*}

\begin{table} [!htbp]
\centering
\begin{tabular}{|c|c|c|c|c|c|c|}
\hline
 & \multicolumn{2}{c|}{Positive} & \multicolumn{2}{c|}{Neutral} & \multicolumn{2}{c|}{Negative} \\
 \hline
 & $M$ & $SD$ & $M$ & $SD$ & $M$ & $SD$ \\
\hline
 \raisebox{-0.1em}{\includegraphics[height=0.9em]{1F60A.png}} & 6.56&0.60 & 5.94&0.92  & 2.92&1.46\\
 \raisebox{-0.1em}{\includegraphics[height=0.9em]{1F60A.png}}$\times$3 & 6.74&0.55 & 6.23&0.85 & 3.14&1.60\\
\hline
 \raisebox{-0.1em}{\includegraphics[height=0.9em]{1F60D.png}} & 6.74&0.58 & 6.09&1.10 & 2.99&1.47 \\
 \raisebox{-0.1em}{\includegraphics[height=0.9em]{1F60D.png}}$\times$3 & 6.78&0.63 & 6.29&0.92 & 3.20&1.55 \\
\hline
 \raisebox{-0.1em}{\includegraphics[height=0.9em]{1F60B.png}} & 6.23&0.85 & 5.90&0.95 & 2.96&1.32\\
 \raisebox{-0.1em}{\includegraphics[height=0.9em]{1F60B.png}}$\times$3& 6.26&0.68 & 6.10&0.96 & 3.01&1.41\\
\hline
 \raisebox{-0.1em}{\includegraphics[height=0.9em]{1F60E.png}} & 6.21&0.80 & 5.53&1.02 & 2.57&1.29 \\
 \raisebox{-0.1em}{\includegraphics[height=0.9em]{1F60E.png}}$\times$3 & 6.36&1.15 & 5.70&1.03 & 2.7&1.33 \\
\hline
 \raisebox{-0.1em}{\includegraphics[height=0.9em]{1F917.png}} & 6.38&0.72 & 5.97&1.16 & 2.91&1.34 \\
 \raisebox{-0.1em}{\includegraphics[height=0.9em]{1F917.png}}$\times$3 & 6.69&0.54 & 6.22&0.83 & 2.83&1.40 \\
\hline
 Control & 6.44&1.10 & 4.25&0.76 & 1.38&0.65 \\
\hline 
\end{tabular}
\caption{Means and standard deviations of sentiment scores of combinations of \textit{\textbf{positive}} emojis and plain verb messages of three valences.}~\label{tab:positiveemojis}
\end{table}

\begin{table} [!htbp]
\centering
\begin{tabular}{|c|c|c|c|c|c|c|}
\hline
 & \multicolumn{2}{c|}{Positive} & \multicolumn{2}{c|}{Neutral} & \multicolumn{2}{c|}{Negative} \\
 \hline
 & $M$ & $SD$ & $M$ & $SD$ & $M$ & $SD$ \\
\hline
 \raisebox{-0.1em}{\includegraphics[height=0.9em]{1F914.png}} & 4.69 & 1.35 & 4.07 & 0.59 & 1.92 & 1.15 \\
 \raisebox{-0.1em}{\includegraphics[height=0.9em]{1F914.png}}$\times$3 & 4.83 & 1.52 & 4.17 & 0.63 & 1.67 & 0.95 \\
\hline
 \raisebox{-0.1em}{\includegraphics[height=0.9em]{1F634.png}} & 4.01 & 1.53 & 3.71 & 0.93 & 1.96 & 0.80\\
 \raisebox{-0.1em}{\includegraphics[height=0.9em]{1F634.png}}$\times$3 & 4.11 & 1.54 & 3.67 & 0.92 & 1.65 & 0.79\\
\hline
 \raisebox{-0.1em}{\includegraphics[height=0.9em]{1F633.png}} & 4.94 & 1.40 & 3.81 & 1.18 & 2.00 & 1.02 \\
 \raisebox{-0.1em}{\includegraphics[height=0.9em]{1F633.png}}$\times$3 & 5.25 & 1.54 & 4.10 & 1.12 & 1.89 & 0.97 \\
\hline
 \raisebox{-0.1em}{\includegraphics[height=0.9em]{1F610.png}} & 4.44 & 1.35 & 3.85 & 0.82  & 1.89 & 0.80\\
 \raisebox{-0.1em}{\includegraphics[height=0.9em]{1F610.png}}$\times$3 & 4.36 & 1.59 & 3.78 & 1.02 & 1.89 & 0.86\\
\hline
 \raisebox{-0.1em}{\includegraphics[height=0.9em]{1F611.png}} & 4.39 & 1.47 & 3.71 & 0.97 & 1.90 & 0.96 \\
 \raisebox{-0.1em}{\includegraphics[height=0.9em]{1F611.png}}$\times$3 & 4.13 & 1.94 & 3.42 & 1.28 & 1.89 & 0.94 \\
\hline
Control & 6.13 & 1.05 & 4.17 & 0.56 & 1.75 & 1.22 \\
\hline 
\end{tabular}
\caption{Means and standard deviations of sentiment scores of combinations of \textit{\textbf{neutral}} emojis and plain verb messages of three valences.}~\label{tab:neutralemojis}
\end{table}

\begin{table} [!htbp]
\centering
\begin{tabular}{|c|c|c|c|c|c|c|}
\hline
& \multicolumn{2}{c|}{Positive} & \multicolumn{2}{c|}{Neutral} & \multicolumn{2}{c|}{Negative} \\
 \hline
 & $M$ & $SD$ & $M$ & $SD$ & $M$ & $SD$ \\
\hline
  \raisebox{-0.1em}{\includegraphics[height=0.9em]{1F621.png}} & 3.08 & 1.85 & 2.24 & 1.04 & 1.63 & 1.07\\
  \raisebox{-0.1em}{\includegraphics[height=0.9em]{1F621.png}}$\times$3 & 2.85 & 2.03 & 1.89 & 1.00 & 1.31 & 0.96\\
\hline
 \raisebox{-0.1em}{\includegraphics[height=0.9em]{1F629.png}} & 3.81 & 1.92 & 2.81 & 1.20 & 1.90 & 1.10 \\
 \raisebox{-0.1em}{\includegraphics[height=0.9em]{1F629.png}}$\times$3 & 3.38 & 1.83 & 2.64 & 1.37 & 1.51 & 0.82 \\
\hline
 \raisebox{-0.1em}{\includegraphics[height=0.9em]{1F62D.png}} & 4.41 & 1.87 & 3.19 & 1.26 & 2.08 & 1.14 \\
 \raisebox{-0.1em}{\includegraphics[height=0.9em]{1F62D.png}}$\times$3 & 4.58 & 2.13 & 2.96 & 1.45 & 1.63 & 1.08 \\
\hline
 \raisebox{-0.1em}{\includegraphics[height=0.9em]{1F614.png}} & 4.24 & 1.56 & 3.38 & 0.99 & 1.90 & 1.16 \\
 \raisebox{-0.1em}{\includegraphics[height=0.9em]{1F614.png}}$\times$3 & 4.04 & 1.65 & 6.29 & 0.92 & 3.20 & 1.55 \\
\hline
 \raisebox{-0.1em}{\includegraphics[height=0.9em]{1F622.png}} & 3.94 & 1.66 & 3.06 & 0.98  & 1.97 & 1.05\\
 \raisebox{-0.1em}{\includegraphics[height=0.9em]{1F622.png}}$\times$3 & 3.94 & 1.92 & 2.89 & 1.26 & 1.47 & 1.00\\
\hline
 Control & 6.26 & 1.14 & 4.20 & 0.69 & 1.81 & 1.25 \\
\hline 
\end{tabular}
\caption{Means and standard deviations of sentiment scores of combinations of \textit{\textbf{negative}} emojis and plain verb messages of three valences.}~\label{tab:negativeemojis}
\end{table}

\begin{table} [!htbp]
\centering
\begin{tabular}{|c|c|c|c|c|c|c|}
\hline
& \multicolumn{2}{c|}{Positive} & \multicolumn{2}{c|}{Neutral} & \multicolumn{2}{c|}{Negative} \\
 \hline
 & $M$ & $SD$ & $M$ & $SD$ & $M$ & $SD$ \\
\hline
 \raisebox{-0.1em}{\includegraphics[height=0.9em]{1F4AF.png}} & 6.24 & 0.73 & 4.90 & 1.03 & 1.80 & 1.02 \\
 \raisebox{-0.1em}{\includegraphics[height=0.9em]{1F4AF.png}}$\times$3 & 6.70 & 0.63 & 5.00 & 1.09 & 1.87 & 1.30 \\
\hline
 \raisebox{-0.1em}{\includegraphics[height=0.9em]{2728.png}} & 6.60 & 0.62 & 4.71 & 0.80 & 1.90 & 0.87 \\
 \raisebox{-0.1em}{\includegraphics[height=0.9em]{2728.png}}$\times$3 & 6.53 & 0.68 & 5.07 & 1.05 & 1.96 & 1.04 \\
\hline
 \raisebox{-0.1em}{\includegraphics[height=0.9em]{1F389.png}} & 6.41 & 0.71 & 4.93 & 0.95 & 2.04 & 1.26\\
 \raisebox{-0.1em}{\includegraphics[height=0.9em]{1F389.png}}$\times$3 & 6.72 & 0.53 & 5.19 & 1.17 & 1.89 & 0.99\\
\hline
 \raisebox{-0.1em}{\includegraphics[height=0.9em]{1F4F7.png}} & 5.89 & 0.97 & 4.11 & 0.43  & 1.47 & 0.98\\
 \raisebox{-0.1em}{\includegraphics[height=0.9em]{1F4F7.png}}$\times$3 & 5.90 & 0.95 & 4.21 & 0.74 & 1.67 & 0.91\\
\hline
 \raisebox{-0.1em}{\includegraphics[height=0.9em]{1F3B6.png}} & 6.59 & 0.63 & 4.76 & 0.86 & 1.64 & 0.72 \\
 \raisebox{-0.1em}{\includegraphics[height=0.9em]{1F3B6.png}}$\times$3 & 6.56 & 0.73 & 4.93 & 0.89 & 1.83 & 1.91 \\
\hline
 Control & 6.11 & 0.89 & 4.00 & 0.71 & 1.56 & 0.86 \\
\hline 
\end{tabular}
\caption{Means and standard deviations of sentiment scores of combinations of \textit{\textbf{non-facial}} emojis and plain verb messages of three valences.}~\label{tab:non-facialemojis}
\end{table}


\end{appendices}

\end{document}